\documentclass[letterpaper]{article} 
\usepackage[]{aaai23}  
\usepackage{times}  
\usepackage{helvet}  
\usepackage{courier}  
\usepackage[hyphens]{url}  
\usepackage{graphicx} 
\usepackage{natbib}  
\usepackage{caption} 
\usepackage{tabularx} 
\usepackage{enumitem} 
\usepackage{array}
\usepackage{booktabs}
\usepackage{float}

\usepackage{algorithm}
\usepackage{algorithmic}

\usepackage{multirow}
\usepackage[table,xcdraw]{xcolor}
\usepackage{hhline}
\usepackage{listings}

\usepackage{newfloat}
\usepackage{listings}
\usepackage[frozencache,cachedir=.]{minted}
\DeclareCaptionStyle{ruled}{labelfont=normalfont,labelsep=colon,strut=off} 
\floatstyle{ruled}
\newfloat{listing}{tb}{lst}{}
\floatname{listing}{Listing}
\pdfoutput=1

\usepackage{todonotes}

\newcolumntype{L}{>{\centering\arraybackslash}m{8cm}}
\newcolumntype{W}{>{\centering\arraybackslash}m{1cm}}
\newcolumntype{M}[1]{>{\centering\arraybackslash}m{#1}}
\newcolumntype{P}[1]{>{\centering\arraybackslash}p{#1}}

\title{Human in the Loop Novelty Generation}
\author{
    Mark Bercasio\textsuperscript{\rm 1}, 
    Allison Wong\textsuperscript{\rm 2}, 
    Dustin Dannenhauer\textsuperscript{\rm 3}
}
\affiliations{
    \textsuperscript{\rm 1}PacMar Technologies, mbercasio@pacmartech.com\\
    \textsuperscript{\rm 2}PacMar Technologies, awong@pacmartech.com\\
    \textsuperscript{\rm 3}Parallax Advanced Research, dustin.dannenhauer@parallaxresearch.com
}

\usepackage{bibentry}

\makeatletter
\def\@copyrightspace{\relax}
\makeatother

\begin{document}

\maketitle

\begin{abstract}

Developing artificial intelligence approaches to overcome novel, unexpected circumstances is a difficult, unsolved problem. One challenge to advancing the state of the art in novelty accommodation is the availability of testing frameworks for evaluating performance against novel situations. Recent novelty generation approaches in domains such as Science Birds and Monopoly leverage human domain expertise during the search to discover new novelties. Such approaches introduce human guidance before novelty generation occurs and yield novelties that can be directly loaded into a simulated environment. We introduce a new approach to novelty generation that uses abstract models of environments (including simulation domains) that do not require domain-dependent human guidance to generate novelties. A key result is a larger, often infinite space of novelties capable of being generated, with the trade-off being a requirement to involve human guidance to select and filter novelties post generation. We describe our Human-in-the-Loop novelty generation process using our open-source novelty generation library to test baseline agents in two domains: Monopoly and VizDoom. Our results shows the Human-in-the-Loop method enables users to develop, implement, test, and revise novelties within 4 hours for both Monopoly and VizDoom domains.  
\end{abstract}

\section{Introduction}

There has been increased interest in artificial intelligence (AI) approaches capable of detecting, characterizing, and accommodating novel situations; situations that violate implicit or explicit assumptions about agents \citep{senator2019science}. AI systems trained to perform tasks in a particular environment rely on large amounts of training data or numerous interactions with an environment simulator. When the environment changes in unexpected ways, these AI approaches typically fails and this is especially true when transitioning an AI system into real-world settings.  

Developing AI systems capable of handling novel situations requires new methods and procedures than those that currently exists. Current effort is spent on getting AI solutions that can solve the original task at hand, achieving robust performance without novelty present. However, it is speculated that a range of additional capabilities are needed in AI systems to accommodate novelty, beyond what is needed to learn to solve a task by itself. 

In most software development practices, tests are created by human users, therefore an automated test-driven development is gaining interest. Also, the verification and validation community has developed processes to obtain requirements of an AI system from the stakeholders of that systems' development in order to understand exactly what the system needs to be able to achieve to be useful. In all these cases, the tests that are used to evaluate an AI's ability come from humans and the task specification encoded in software. These test cases are known ahead of time.  We have clear approaches to testing AI systems in this manner.



Current approaches uses procedural generation to create novelty with the help of a human expert experienced in the domain. This approach have a key advantage in generating more scenarios compared to hand written test cases by human.  They remain limited because they are guided primarily from human expert knowledge about the domain, and thus the space of novelties they can generate are limited and biased within the scope of the human. Humans generally have good intuition about bad novelties, defined as irrelevant, noticeable, or controllable \cite{dannenhauertransforming}. By having a human expert build targeted, automatic scenario generators for domains in which they have expert knowledge helps avoid generating bad or irrelevant novelties.

We developed a new human in the loop novelty generation process that leverages human domain experts intuition about what makes a good novelty with a new automatic novelty generation process that improves on the efficiency of previous solutions. We briefly introduce our novelty generator in the next section. Interested readers may refer to \citep{molineaux2023framework} for more details on that system. Additionally our novelty generator is available on Github\footnote{https://github.com/Parallax-Advanced-Research/noveltygen}. 

The primary contributions of this paper includes:

\begin{enumerate}
    \item A 6-step process by which a human domain expert with developer skills can leverage our novelty generator tool to save time in generating novelties, and consider a greater range of novelties with a reduced bias from human imagination.
    \item An experimental setup showing that novelties generated in this way are capable of challenging agents across multiple environments.
    \item Experimental results from two domains: Monopoly and VizDoom
    \item A qualitative review of the experience for human developers to use our novelty generator.
\end{enumerate}

Section 2 describes our approach in the context of other similar novelty generation approaches. In Section 3 we briefly cover our novelty generation tool, including how to run it, what the output looks like, etc. In section 4 we describe the human-in-the-loop process where a human uses the novelty generator. In Section 5 we describe our experimental setup where we used the novelty generator to generate novelties in two domains: Monopoly and Vizdoom. Section 6 presents results from the empirical evaluations of the novelties while Section 7 describes a qualitative review of the pros and cons of the human in the loop process from our developers who followed the process to generate the novelties described in Sections 5 and 6. Finally we conclude the paper with a discussion for future work.


\section{Related Work}

Origins of automatic novel environment generators can be traced back to Metagame \citep{pell1992metagame} and EGGG \citep{orwant2000eggg}.  Both were designed to find novel variations of chess-like games. More recent work in game design research decoupled the design of a game space from the exploration of that space to find interesting games \citep{smith2010variations}. In their work, the language used to design game spaces was large and flexible compared to the prior chess-like games, but still incorporated many assumptions, such as the existence of a rectangular play space. Even newer work on algorithmic procedural content generation for games focuses on bounded spaces such as levels, characters and quests \cite{risi2020increasing}. In these systems some domain knowledge is encoded by humans to guide the search towards producing valuable artifacts.

Current benchmark domains to evaluate AI agent's novelty-handling capability are designed in close alignment with simulated environments. NovelGridworlds \cite{goel2021novelgridworlds} is an OpenAI Gym environment that leverages human guidance on grid-based environments. Another grid-based novelty generator uses an ontology of novelties related to sequential decision making, such as distinguishing between object and action novelties \cite{balloch2022novgrid}. Birds (an Angry Birds AI domain; \cite{xue2022science}) implements novelties as modifications to the C\# codebase. This includes changing physical parameters and colours of existing game objects to more difficult modifications such as introducing a new class of objects (e.g. hostile external agents that hinders the agent).  GNOME (Generating Novelty in Open-World Multi-agent Environments; \cite{kejriwal2021multi}) is a simulation platform developed to test AI agent's response to novelty in the classic Monopoly board game, using a library of novelty-
generating methods.  An example of this is replacing existing game event functions (like a player passing go and collecting \$200) with human inspired function. Functions maintain a similar signature enabling them to be easily swapped out with the original function (such as a function that takes a wealth tax from the player).




\section{Novelty Generation Tool}
Our approach to novelty generation is originally inspired by Wiggins' \citeyear{wiggins2006preliminary} paper on the Creative Systems Framework. Novelties are created using two types of transformations: R-transformations that change the space of possible states and state transitions, and T-transformations that modify the search method for generating starting states in that environment.

R-transformations\footnote{R and T refer to Wiggins' Creative Systems Framework theory rule sets, otherwise 'R' and 'T' are simply variable names.} modify a domain file describing the state-transition system of an environment. This includes possible states with symbolic constructs like object type hierarchies, relations between object types, object properties, and fluents. Changes to these structures in effect changes the state space. Domains also define the set of state-transitions that are valid in an environment using action, event, and process models. Changes to these models changes the transition space between states. All changes of this nature (either to the state space or transition space) fall under R-transformations.

Our environment-transformation based novelty generator that we present here is based on domain-independent environments models. Using a formal language inspired from the AI Planning community, Planning Definition Domain language (PDDL), we can generate novelties for any domain capable of being modelled with the language. For the full list of constructs that our Transformation Simulation Abstraction Language (TSAL) supports, refer to \cite{dannenhauertransforming}. A key benefit of our novelty generator is that it is domain-independent since it can generate novelties for any environment modeled in the TSAL language; however the domain-independent novelties it produces needs to be guided. We have proposed domain independent heuristics for filtering novelties on three conditions:

\begin{description}
    \item[Relevant:] The novelty affects the agents performance on the task.
    \item[Noticeable:] The novelty causes different observations to the agent.
    \item[Controllable:] The novelty rewards agent policies differently.
\end{description} 

In \cite{dannenhauertransforming} we show that using a simple re-planning agent and a simulated version of a TSAL environment, novelties can be automatically filtered out based on relevance, with the rest left for future work.

Our novelty generator is implemented in python, available on GitHub, and comes with three domains already modeled in the TSAL language.  Two of these domains are described in Section 5.

The caveat with having a general, domain-independent novelty generator is ensuring only "high-quality" or "interesting" novelties are found. To alleviate this problem we introduce the approach in this paper where a human software developer works alongside the novelty generator. The effect is that the human is able to spend less time on the creative brainstorming phase and instead focus their energy on the "evaluation" phase. This also reduces the bias from only producing novelties along certain dimensions (such as the human getting fixated on novel kinds of objects instead of action/event properties). 

\begin{listing}[ht]
\begin{minted}{python}
1 from noveltygen.novelty_generator import [...] as NG
2 domain_fn='domains/blocksworld/domain.tsal'
4 ng = NG(domain_file=domain_fn)
3
4 # choose your RTransformations 
5 r_trans = [RTransformation.ADD_EFFECT,
6            RTransformation.ADD_PRECONDITION,
7            RTransformation.REMOVE_EFFECT]
8      # ... see RTransformation.py for full list
9 
10 # generate a novel environment
11 transform = ng.gen_r_transform(transformations=r_trans)

\end{minted}

\caption{Python example to run Novelty Generator}
\label{listing:1}
\end{listing}

\section{The Human-Dev ITL Process}
The HITL method is time-efficient and removes human bias when developing novelties. Current method of novelty generation requires humans to perform brainstorming sessions to develop novel situations; and such an approach is highly time intensive, bias, and risks rework. Limitations of the current method are mitigated with the proposed HITL process that enables the user to generate, implement, and test novelties within a short time frame relative to the current typical method.

A general overview of our proposed novelty generation process is detailed below. Depending on the domain and expertise, the time to complete these steps varies.
\begin{description}
    \item[Step 1: Construct domain specific TSAL files] \leavevmode \\
    The novelty generator requires a TSAL domain and problem file as input for novelty generation. Time to complete this initial step is variable depending on the HITL’s skill at creating an abstraction of the target domain and their ability to debug issues when passing the domain file through the TSAL interpreter.
    \item[Step 2: Run novelty generator using TSAL domain file] \leavevmode \\
    Once the TSAL domain file has been completed, the HITL will need to edit the novelty generator parameters to target a specific novelty type. We’ve found that a minimum of around 100 generated novelties is enough for further parsing.
    \item[Step 3: Identify possible novelties from generated files] \leavevmode \\
    The HITL must now manually look through the generated files and pick a novelty they think is viable. If the HITL sees fit, they may do some post-processing of certain actions or events by selecting noteworthy pre-conditions or events that can be moved from other generated files.
    \item[Step 4: Implement the novelty] \leavevmode \\
    The next step is to implement the novelty into the domain’s simulation space. Time to completion is variable depending on the modularity of the codebase and the complexity of the novelty. Initial implementation for specific novelty types or levels may take longer in order to set up the initial injection architecture, whereas subsequent implementations are relatively easier and quicker to complete.
    \item[Step 5: Test the novelty] \leavevmode \\
    The fifth step is to run experiments in order to test the viability of the novelty. These experiments consist of running a baseline agent through pre- and post-novelty scenarios while collecting data on the results. The time it takes to complete this step varies on the computation speed of the machine and the number of instances an experiment has. Viable novelties are determined by a significant increase or decrease in baseline agent performance when compared to a pre-novelty scenario.
    \item[Step 6: Revise the novelty] \leavevmode \\
    The final step is to revise novelty based on the experimental results. If the novelty does not have a significant increase or decrease in baseline agent performance when compared to a pre-novelty scenario, the HITL may revise the novelty generator parameters or parameters in the generated files to make the novelty viable. We consider this step optional as some novelties may never be viable, even with revisions.
    
\end{description}

\section{Experimental Setup}
To evaluate our proposed novelty generation process, a series of experiments was conducted in two domains to investigate the limitations and capabilities of the process.
The two domains used to evaluate the HITL novelty generation process were (1) Monopoly and (2) VizDoom. The experimental setup for each domain are detailed below.

\subsection{Monopoly Domain}

The Monopoly domain is a python program that simulates the Monopoly board game that is developed under ISI's GNOME\footnote{https://github.com/mayankkejriwal/GNOME-p3}, an experimental platform created to evaluate the performance of multi-agent AI systems in the face of novelty. GNOME separates the development of AI game-playing agents from the simulator, allowing for the creation and injection of unanticipated novelty not disclosed to the agent.

\subsubsection{Agent Setup} 
Agents used in this domain can be found in the GNOME-p3 GitHub repository with the following modified rule set:
        \begin{itemize}
            \item Players don’t receive an extra turn on doubles
            \item Players automatically set free from jail after one turn
        \end{itemize}

The experiments consisted of four total agents: one baseline reinforcement learning (RL) agent (Agent 1) and three identical background agents (Agent 2, Agent 3, and Agent 4). All three of these background agents follow the same decision-making processes in order to win the game. The background agents achieve victory through proper trades and property management in order to bankrupt the rest of the players. They are not biased against the player agent in the sense that they do not make decisions that are detrimental to ONLY the player agent.

\subsubsection{Evaluation Setup} 
To evaluate the HITL novelty generation process, each of the generated novelties were evaluated by their viability, or impact on the baseline agent.  This was done by comparing win rates of agents in pre- and post-novelty tournaments.  The average wins and standard deviation of the pre-novelty agent performance is shown in Table 1.
\begin{table}[H]
    \begin{center}
    \caption{Results of pre-novelty agent performance over 20 seeds of 10,000 games each.}
    \label{tab:my-table}
    \resizebox{\columnwidth}{!}{
        \begin{tabular}{ c | c | c } \hline
        \rowcolor[HTML]{000000}
        \multicolumn{3}{|c|}{{\color[HTML]{FFFFFF} \textbf{Standard Deviation of Pre-Novelty Agent Performance}}} \\ \hline
         & \textbf{Agent 1} & \textbf{Default Background Agent} \\ 
        \hhline{=|=|=}
        Average Wins         & 6943 & 1019 \\ \hline
        Raw Std Dev          & 45.66 & 30.07 \\ \hline
        \% Win Rate Std Dev  & 0.66  & 2.96 \\ \hline
        \bottomrule
        \end{tabular}
        }
    \end{center}
\end{table}

The three levels of viability used to evaluate the novelties are defined below:

\begin{description}
    \item[Low Viability:] 0.66\% $<$ Agent 1 $<$ 4\% 
    \item[Medium Viability:] 4\% $<$ Agent 1 $<$ 9\%  
    \item[High Viability:] Agent 1 $>$ 9\%
\end{description} 

The change in win rate used to assess viability of a novelty could either be an increase or decrease in win rate. In other words, a novelty that alters the baseline agent’s performance significantly enough could be considered viable.

For each experiment, we simulated three tournaments with 10,000 instances (games) per tournament. To keep results consistent, we used meta seeds 5, 10, 999 as inputs for the tournaments. The novelties developed using the HITL method for each novelty level are summarized in Table 2.


\begin{table}[H]
    \begin{center}
    \caption{Summary of developed agents used to verify and validate generated novelties of various levels.}
    \label{tab:my-table}
    \resizebox{\columnwidth}{!}{
        \begin{tabular}{ c | m{6cm} } \hline
        \rowcolor[HTML]{000000}
        \multicolumn{2}{|c|}{{\color[HTML]{FFFFFF} \textbf{Experiment Information}}} \\ \hline
        \textbf{ID} & \textbf{Description} \\ \hhline{=|=}
        0 & Players gain \$1,000 when passing GO \\ \cline{1-2}
        1 & Players lose \$500 when passing GO \\ \cline{1-2}
        2 & Players can only move when dice values are identical \\ \cline{1-2}
        3 & Players gain \$25 at the end of their turn \\ \hline 
        4 & Player must be in jail to receive a property after trade is complete \\ \cline{1-2} 
        5 & Player must be in jail to receive cash after trade is complete \\ \cline{1-2}
        6 & Property is automatically sold back to the bank after rent is paid for that property \\ \cline{1-2}
        7 & Rent is not paid unless the owner of the space is in jail \\ \cline{1-2}
        8 & The player cannot collect money if they are the last bidder to an auction \\ \hline 
        \bottomrule
        \end{tabular}
        }
    \end{center}
\end{table}

\subsection{VizDoom Domain}
VizDoom is an open-source project used as a research platform for RL using the popular first-person shooter game, Doom (1993), as the environment. This project allows researchers and developers to create and test learning and decision-making algorithms through the modular simulation environment. We developed on top of  Washington State University's (WSU) VizDoom Novelty Generator GitHub\footnote{https://github.com/holderlb/WSU-SAILON-NG/tree/master/WSU-Portable-Generator} to inject our custom novelties and conduct experiments on RL agents.

\subsubsection{Agent Setup} 
WSU provided their state-of-the-art (SOTA) player agent for testing, which we used as our baseline agent. While we are unable to go over the specifics of the baseline player agent's RL logic, we are able provide a general overview of its functionality. 

The agent has eight actions: forward, backward, left, right, turn left 45 degrees, turn right 45 degrees, shoot, and no action. For each turn, a feature vector of the current state of the simulation environment is sent to the agent; including general information about enemies, items, and the player agent. In response, the agent provides an action to be performed. Performance is defined as the amount of time left in the episode divided by the maximum time for the episode. 

Training for the agent is done separately from testing. The user is free to indicate the number of training instances the agent should be trained on. When training is complete, the current model for the agent is saved in memory. We have pre-trained the SOTA agent for 100 pre-novelty instances before using it for novelty testing.

\subsubsection{Evaluation Setup}
The VizDoom domain consists of a fork of the WSU VizDoom GitHub\footnote{https://github.com/mbercasio94/WSU-SAILON-NG}; with a change of four total enemies and four of each item (health, ammo, trap) spawned for every instance. We conducted one experiment with 1000 instances for pre-novelty, and 1000 instances of each post-novelty. As a control and point of comparison, the SOTA agent was able to win 991 out of 1000 pre-novelty games averaging to a win-rate of 99.1 percent.


The novelties developed using the HITL method for each novelty level are summarized in Table 3.
\begin{table}
    \begin{center}
    \caption{Summary of developed agents used to verify and validate generated novelties of various levels.}
    \label{tab:my-table}
    \resizebox{\columnwidth}{!}{
        \begin{tabular}{ c | m{6cm} } \hline
        \rowcolor[HTML]{000000}
        \multicolumn{2}{|c|}{{\color[HTML]{FFFFFF} \textbf{Experiment Information}}} \\ \hline
        \textbf{ID} & \textbf{Description} \\ \hhline{=|=}
        0 & Player loses 1 health point every 7 turns \\ \cline{1-2}
        1 & Player loses one ammo clip every 40 turns \\ \cline{1-2}
        2 & Player is turned 90 degrees every 3 turns \\ \hline 
        3 & After rotating, the player loses 2 health points \\ \cline{1-2} 
        4 & After shifting positions, the player loses 1 health point \\ \cline{1-2} 
        5 & After rotating, the player is turned 15 degrees to the left \\ \hline
        \bottomrule
        \end{tabular}
        }
    \end{center}
\end{table}

\section{Results}

\subsection{Novelty Generation Setup}
Independent of the TSAL file creation explained in step 1 of the HITL novelty creation process, the total time to find, implement, and test a new novelty ranges from 2 hours to 4 hours.  Step 1 is removed from the time band since it varies depending on the user expertise and the target domain's complexity.

\subsection{Monopoly}
The performance results of Agent 1 for all conducted novelties is shown in Table \ref{tab:monopoly results}. The results indicate the majority of implemented novelties are of high viability, meaning a significant difference of $>$ 9\% in the agent performance. Table 4 contains a summary of the agent's change in win rate over 10,000 games for each novelty.

\begin{table}[H]
    \begin{center}
    \caption{Viability of generated Monopoly novelties}
    \label{tab:monopoly results}
    \resizebox{\columnwidth}{!}
        {
        \begin{tabular}{ c | c | c | c | c  } \hline
        \rowcolor[HTML]{000000}
        \multicolumn{5}{|c|}{{\color[HTML]{FFFFFF} \textbf{Novelty Performance Results}}} \\ \hline
        \multicolumn{1}{c}{} & \multicolumn{3}{|c|}{\textbf{Post-Novelty Agent Absolute Win Rate}} & \multicolumn{1}{c}{} \\ \cline{2-4}
        \textbf{ID} & Seed = 5 & Seed = 10 & Seed = 999 & \textbf{Viability} \\ \hhline{ =|=|=|=|= }
        0  & -20.21\% & -19.51\% &  -22.05\% &  \textbf{High} \\ \hline
        1  &   5.14\% &   4.55\% &    4.04\% &  Medium \\ \hline
        2  & -28.61\% &	-27.11\% &	-26.90\% &	\textbf{High} \\ \hline
        3  & -26.71\% &	-26.04\% &	-28.01\% &	\textbf{High} \\ \hline
        4  & -20.09\% &	-19.35\% &	-18.97\% &	\textbf{High} \\ \hline
        5  & -15.59\% &	-15.17\% &	-16.31\% &	\textbf{High} \\ \hline
        6  & -28.06\% &	-30.33\% &	-28.42\% &	\textbf{High} \\ \hline
        7  & -99.90\% &	-99.87\% &	-99.94\% &	\textbf{High} \\ \hline
        8  & -96.48\% &	-95.99\% &	-96.25\% &	\textbf{High} \\ \hline
        9  &  15.87\% &	 14.75\% &	 14.41\% &	\textbf{High} \\ \hline
        10 & -49.54\% &	-48.39\% &	-48.48\% &	\textbf{High} \\ \hline
        11 & -21.89\% &	-20.41\% &	-20.26\% &	\textbf{High} \\ \hline
        \bottomrule
        \end{tabular}
        }
    \end{center}
\end{table}

\subsection{VizDoom}
The performance results of the SOTA agent for all conducted novelties is shown in Table 7. Since the average SOTA pre-novelty win rate was 99.1\%, any novelties below a 90\% win rate were considered high viability. Table 5 contains a summary of the agent's win rate over 1,000 games for each novelty.

\begin{table}[h]
    \begin{center}
    \caption{Viability of generated VizDoom novelties.}
    \label{tab:my-table}
    \begin{tabular}{|c|c|c|} \hline
    \rowcolor[HTML]{000000}
    \textcolor{white}{\textbf{ID}} & \textcolor{white}{\textbf{Score \%}} & \textcolor{white}{\textbf{Viability}} \\ \hline
    0  &   52.6\% &   \textbf{High} \\ \hline
    1  &   46.0\% &   \textbf{High} \\ \hline
    2  &   39.8\% &   \textbf{High} \\ \hline
    3  &    0.4\% &   \textbf{High} \\ \hline
    4  &   31.5\% &   \textbf{High} \\ \hline
    5  &    0.0\% &   \textbf{High} \\ \hline
    \bottomrule
    \end{tabular}
    \end{center}
\end{table}

\subsection{Discussion}
During our experiments, most novelties using the HITL novelty generation process resulted in medium to high viability novelties, reinforcing our hypothesis that the generated novelties are typically sufficient to thoroughly test various agents. In the situation where a generated novelty was categorized as low viability, we were able to regenerate, re-implement, and retest new novelties in less than 4 hours. Our novel HITL method is an automated and time-efficient novelty generation process compared to current human-only novelty generation process.  

\begin{table*}[htbp]
    \begin{center}
    \caption{Novelty Generator Experiment Results with HITL completion time}
    \label{tab:novelty-results}
        \begin{tabular}{|p{0.5\textwidth}|>{\centering\arraybackslash}p{0.2\textwidth}|>{\centering\arraybackslash}p{0.2\textwidth}|} \hline
        \rowcolor[HTML]{000000}
        \multicolumn{3}{|c|}{{\color[HTML]{FFFFFF} \textbf{Approximate Time to Generate Novelties per Step while Working with the Novelty Generator}}} \\ \hline
        \textbf{HITL Novelty Generation Step Description} & \multicolumn{2}{c|}{\textbf{Avg. Minutes to Complete Step}} \\ \hhline{=|=|=}
        & \textbf{VizDoom} & \textbf{Monopoly} \\ \hline
        \textbf{Step 1: Construct TSAL file} specific to target domain & Variable & Variable \\ \hline
        \textbf{Step 2: Run novelty generator using TSAL domain file.} This process includes editing of the novelty generator parameters to target specific novelty levels. Novelty file generation - target to generate at least 100 files. & 5 minutes & 5 minutes \\ \hline
        \textbf{Step 3: Identify possible novelties from generated files.} Manually parse through generated files for novelties that may prove viable for the target novelty level. Post-processing of certain conditions and effects sometimes required. & ~30 minutes per novelty & ~30 minutes per novelty \\ \hline
        \textbf{Step 4: Implement the novelty.} Time to implement a novelty depends on the level of that novelty. Initial implementation for a specific novelty  level is slow (requires planning of software architecture). Subsequent implementations significantly faster for already created novelty levels. & 45 minutes for the first novelty, 20 minutes for subsequent novelties  & 60 minutes for the first novelty, 20 minutes for subsequent novelties \\ \hline
        \textbf{Step 5: Test the novelty.} Run experiment to test for novelty viability. Dependent on computation speed of the simulation machine. & ~30-60 minutes & ~1-2 hours \\ \hline
        \textbf{Step 6: Revise the novelty.} Manually revise novelty values as needed. May also discard novelty if no changes work. & ~5 minutes & ~5 minutes \\ \hline
        \bottomrule
        \end{tabular}
    \end{center}
\end{table*}

\section{Qualitative Review of Human Experience during Novelty Generation}
In this section we go over a HITL’s initial thoughts for each step of the novelty generation process. This provides a look at the possible advantages and limitations of our current model in the perspective of a first-time user. Table 6 provides an overview of the time it takes to complete each step for both Monopoly and VizDoom domains.

\begin{description}

\item[Step 1: Construct domain specific TSAL files] \leavevmode \\
 \textbf{(Difficulty - High)}
For those unfamiliar with PDDL or declarative languages in general, it might take several iterations and possible feedback from an expert in order to properly model the domain. Abstraction of a domain must account for certain nuances that might not be so obvious on the initial iteration. 

In the case of the Monopoly domain, an extra turn from dice rolls was modelled in the TSAL file even though it was not part of the actual simulation logic, thus resulting in extra turn novelties that deviated too much from the original domain.
\item[Step 2: Run novelty generator using TSAL domain file] \leavevmode \\
 \textbf{(Difficulty - Low)}
Slight training by an expert or readme is required in order to know which lines of novelty generator code need to be adjusted in order to target specific novelty types.  
\item[Step 3: Identify possible novelties from generated files] \leavevmode \\
\textbf{(Difficulty - Low)}
As the HITL becomes more familiar with the domain, they should have a better understanding of the novelties that are likely to cause significant impact on agent performance. For example, the HITL may find a pattern where adjusting the preconditions of a specific event has higher likelihood of producing a viable novelty.
\item[Step 4: Implement the novelty] \leavevmode \\
\textbf{(Difficulty - Medium)}
Creating novelty injection code depends on the modularity of the domain – how easy is it for a developer to adjust the rules and environment of the simulation space? In addition, difficulty depends on the type of novelty being implemented. For example, novelties that effect agent goals may require further implementation work when compared to a novelty that simply changes an event-based rule of the world. Similar to step 3, as the HITL familiarizes themselves with the codebase, implementation of further novelties should be relatively easier.
\item[Step 5: Test the novelty] \leavevmode \\
\textbf{(Difficulty - Low)}
In terms of manual labor, a HITL will need to simply create a batch script in order to run experiments on the implemented novelties. Running experiments with more instances will take longer to complete, but results in a more accurate view of novelty viability. The threshold for the minimum change in performance required for a novelty to be deemed viable will depend on the judgement of the HITL.
\item[Step 6: Revise the novelty] \leavevmode \\
\textbf{(Difficulty - Low)}
A HITL will need to simply change parameter(s) in the generated novelty file or novelty generator. As an example, for \textit{Experiment 0} (Players gain \$1000 when passing GO), the value of \$1000 chosen by the novelty generator may not cause a significant change in agent performance. A user may opt to manually change the value to \$500 which may end up resulting in a more significant change in agent performance. Of course, some novelties will never be viable even with HITL revision. In that case, it is safe to discard the novelty and try a new one.

\end{description}

\section{Conclusions and Future Work}
We introduced a 6-step process by which a human with domain expertise can work alongside our open-source novelty generator to discover, implement, and score AI approaches. Our novelty generator is domain-independent, as demonstrated by our evaluation in two substantially different domains: Monopoly and VizDoom. Any environment that is defined in the Planning Domain Definition Language (PDDL) can be easily modified to work with our novelty generator, since our TSAL language includes most PDDL constructs.

We hope to apply our novelty generator to more environments and develop a more complete testbed environment where agents can be tested directly against the state-transition system without requiring human developer effort to implement novelties in an external simulator. The TSAL language is capable and existing PDDL simulators exist, however at this time a TSAL simulator has not yet been built. A TSAL simulator would enable novelties produced by our method to be used in other frameworks like PDDL-gym \citep{silver2020pddlgym}: a testbed for reinforcement learning approaches in PDDL domains. 

Additionally, leveraging human guidance would be easier with a better user interface. Currently, our system writes novelties to a plain text file that a human can scan line-by-line. An improved user interface should offer features such as syntax highlighting and other visual indicators to aid the human in picking out good novelties, which we leave for future work.

While the results demonstrated our generator's effectiveness in these areas, a significant limitation arises from its restricted scope. Our current model may not generalize well to other domains or interdisciplinary contexts. By limiting the generator's capabilities to only two domains and two novelty levels, we may have inadvertently curtailed its potential to create groundbreaking ideas across a broader range of disciplines. Future research should focus on expanding the generator's reach to encompass multiple domains, thus allowing for a more comprehensive exploration of novel concepts and enhancing its applicability in various fields.

\bigskip
\noindent \textbf{Acknowledgements:} This work was supported by the Defense Advanced Research Projects Agency (DARPA) under Contract No. HR001121C0236. Any opinions, findings, and conclusions or recommendations expressed in this material are those of the author(s) and do not necessarily reflect the views of the Defense Advanced Research Projects Agency (DARPA).

\bibliography{aaai23}

\end{document}